\documentclass[9pt]{extarticle}
\usepackage{spconf,amsmath,graphicx}
\usepackage{booktabs}
\usepackage{tikz}

\usepackage{caption}
\usepackage{subcaption}

\title{Contextual Biasing of Named-Entities with Large Language Models}
\name{Chuanneng Sun$^{1*}$\thanks{*The first author performed the work as a research scientist intern at Meta AI.}, Zeeshan Ahmed$^2$, Yingyi Ma$^2$, Zhe Liu$^2$, Lucas Kabela$^2$, Yutong Pang$^2$, Ozlem Kalinli$^2$}
\address{$^1$Rutgers University--New Brunswick, Piscataway, NJ, USA\\
$^2$Meta AI, Menlo Park, CA, USA}
\begin{document}
%
\maketitle
\begin{abstract}
We explore contextual biasing with Large Language Models (LLMs) to enhance Automatic Speech Recognition (ASR) in second-pass rescoring.
Our approach introduces the utilization of prompts for LLMs during rescoring without the need for fine-tuning. These prompts incorporate a biasing list and a set of few-shot examples, serving as supplementary sources of information when evaluating the hypothesis score. Furthermore, we introduce multi-task training for LLMs to predict entity class and the subsequent token. To address sequence length constraints and improve the efficiency of contextual biasing, we propose dynamic prompting based on class tag predictions. Through dynamic prompting, we leverage the class tag predictions to identify the most probable entity class and subsequently utilize entities within this class as biasing context for the next token prediction. We evaluate the performance of proposed methods in terms of Word Error Rate (WER) on an internal entity-heavy and the SLUE-Voxpopuli datasets. Our results show significant improvements: biasing lists and few-shot examples achieved a relative improvement of 17.8\% and 9.6\%, while multi-task training and dynamic prompting achieved 20.0\% and 11.3\% relative WER improvement, respectively.
\end{abstract}
\section{Introduction}
Language Models (LMs) in Automatic Speech Recognition~(ASR) systems suffer from a drastic reduction in quality when recognizing uncommon words, e.g., Named Entities (NE) that appear infrequently
in training data~\cite{zhao2019shallow, le2021deep, le21_interspeech, ma2023adaptive}.
These words are highly personalized and thus hard to be accommodated in a unified model that fits all users.
Generally, these uncommon words can be inferred from the context in which the model is being used.
For example, speech assistants in the calling and messaging domain employ the contact list to enhance name recognition by favoring recognition associated with the contact list.

Various methods have previously been employed to incorporate contextual information into the model. In~\cite{7178957, 1326104}, an approach is presented to improve the contact name recognition using n-gram based class LM~\cite{brown1992class} and dynamic weighted finite state transducer (WFST). The WFST, i.e., the user's contact model, models relevant names from the user's contact list. Subsequently, a class LM is then trained by replacing the names in the data with a special token \$CONTACT. During inference, \$CONTACT is replaced by the user’s contact model. The finite state replacement mechanism allows for this on-demand substitution.
This method has long been employed to inject domain-specific contextual information into the system.
A similar kind of biasing can be applied to Neural Network Language Models (NNLMs)~\cite{9747573} where a class NNLM is used instead of a n-gram class LM. The class NNLM is used in shallow fusion with the ASR system during beam-search decoding. Instead of an FST-replace with the user’s contact model, the biasing is applied dynamically when the class NNLM predicts any of the class tokens
e.g., \$CONTACT.
Both the class LM and NNLM described above have improved the recognition of entities during shallow fusion with the ASR system with contextual biasing.

Neural networks have demonstrated adaptability in directly incorporating contextual information into their architecture. In previous studies~\cite{9383560, 9687895}, researchers have illustrated how contextual information, presented as an embedding vector, can effectively bias predictions within the acoustic network.  Both these two methods use a special-purpose encoder to encode the contextual knowledge.
There are also several works investigating personalized LMs to improve the ability of LMs when working with rare words.
\cite{9383560} uses a Personalized LM~(PLM) predictor network to produce an embedding vector that takes the previous token history and a list of contextual items.
In \cite{salemi2023lamp}, the authors proposed a personalization framework for working with LLMs. Particularly, according to each user's profile, a subset of words will be selected from the user's biasing list. However, this method cannot work with utterances that have little context (e.g., short utterances). In \cite{fu2023robust}, the authors proposed a personalized conformer transducer-based ASR model with a pretrained NNLM to encode the utterance and biasing entities. The biasing is performed in both the acoustic and the language domains.
On the other hand, 
attention-based contextual biasing methods have also been explored~\cite{9687895, pundak2018deep, chen2019joint, jain2020contextual}.
In this kind of method, an encoder network is used to encode a list of contextual items, and a Multi-Head Attention (MHA) biasing layer is used to attend to the context for each audio frame. However, these entities are often not correlated with each other, indicating that compressing them into embeddings will cause a loss of information.
Recently, there have been works exploring the use of LLMs in ASR systems~\cite{fathullah2023prompting, 10096429, udagawa2022effect}. \cite{10096429, udagawa2022effect} studied the sentence scoring method using LLM in the ASR system while, in \cite{fathullah2023prompting}, an audio-domain biasing approach is proposed. Although these methods have taken advantage of the ability of LLMs, their methods do not incorporate additional contextual information.

The aforementioned traditional methods are not flexible enough to work for new classes while the neural network-based methods cause information loss when compressing contextual information because entities like person names are not correlated. To reconcile these problems, we propose a Large Language Model (LLM)-based contextual biasing method and two learning methods--few-shot prompt learning and multi-task training, to improve performance.

Our contribution can be summarized as follows:
\begin{enumerate}
    \item We propose contextual biasing prompts when calculating the second-pass score. We propose a simple yet effective format of prompts to incorporate the lists of biasing entities. Moreover, we propose a few-shot learning method by providing examples in the prompts.
    \item To further enhance performance, we introduce a multi-task training framework. This involves augmenting the Large Language Model (LLM) with a dedicated tag head that predicts the entity class tag (e.g., person, location) of the next tokens. The entire model is jointly trained with two distinct losses: the entity tag prediction loss and the token prediction loss. To integrate class tag prediction into token prediction, we employ the Gumbel softmax trick, facilitating gradient propagation.
    \item We propose dynamic prompting to improve attention efficiency and to reduce the input sequence length. For each token, we first obtain the entity class prediction from the class head (top 1). Then, we subset the contextual information in the prompt based on the class for token prediction.
\end{enumerate}

\section{Proposed Work}

In harnessing the capacity of Large Language Models (LLMs) to incorporate supplementary information as input, we put forth a novel approach to infusing contextual information into prompts. This augmentation of prompts serves to steer the model's focus and bias. Besides prompts, we propose multi-task training and dynamic prompting to further improve the performance.

\vspace{-.05in}

\subsection{Biasing via Prompts}
\begin{figure}[t!]
    \vspace{0.1in}
    \centering 
    \begin{subfigure}[b]{0.45\textwidth}
        \centering
        \includegraphics[width=1\textwidth]{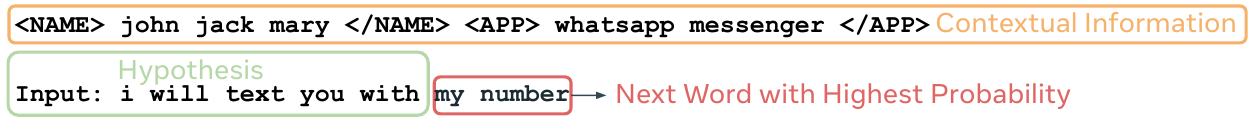}
        \caption{\label{fig:prompt_no_example} Providing contextual information only to the LLM.}
    \end{subfigure}\hfil
    \begin{subfigure}[b]{0.45\textwidth}
        \centering 
        \includegraphics[width=1\textwidth]{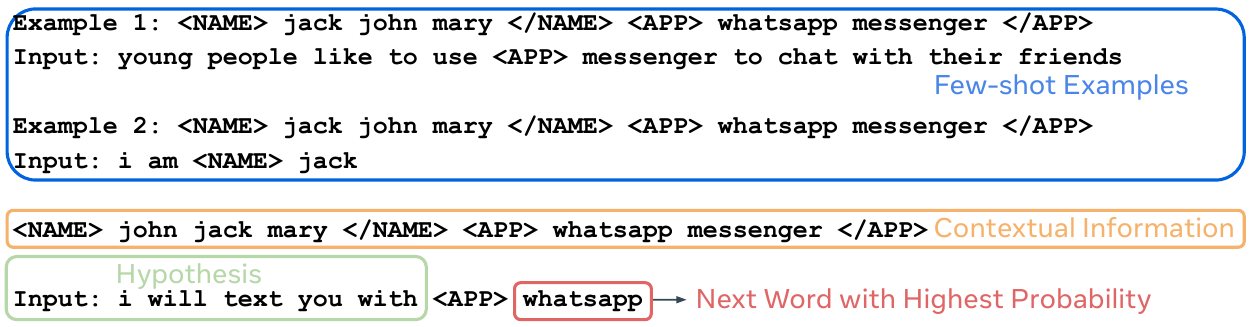}
        \caption{\label{fig:prompt_example} Providing few-shot examples, entity class tags, and contextual information to the LLM.}
    \end{subfigure}
    \caption{\label{fig:prompt_demo} Demonstration of the power of prompts in contextual biasing. Given two-shot example and the entity class tag (e.g., $<$NAME$>$), the LLaMA model is able to assign a high probability to correct words.}
    \vspace{-.1in}
\end{figure}

Upon observing the remarkable capabilities demonstrated by Large Language Models (LLMs) such as ChatGPT~\cite{openai_2022} and LLaMA~\cite{touvron2023llama}, the notion of achieving contextual biasing through the augmentation of prompts with supplemental information appears promising.
Figure~\ref{fig:prompt_demo} demonstrates a simple yet illustrative example of the power of prompting with LLaMA.
We can observe that few-shot examples can help improve the quality of the generated words. Furthermore, the tags (e.g., $<$NAME$>$) are also helpful for the model to comprehend the context, which motivates us to propose the multi-task LLM for entity class prediction.
Although the use of prompts seems straightforward, the intricacies of prompt design are far from trivial. The challenge lies in effectively infusing contextual cues without introducing a surplus of redundant tokens, such as punctuation.
To strike this balance, we adopt the subsequent format for our prompts: \textit{$<$Entity Class$>$Entities$<$/Entity Class$>$}. This simplistic structure is chosen to avoid surpassing the maximum sequence length constraint, considering the likelihood of numerous entities within the data.

Furthermore, we explore the few-shot prompt learning strategy, enhancing the model's performance by incorporating illustrative instances directly into the prompts. Our prompts adopt the following format: \textit{Example 1: Example 2. $<$Entity Class$>$Entities$<$/Entity Class$>$ Input: Input Sentence}. Each example encapsulates pertinent entity information and its corresponding input sentence, randomly sampled from the dataset.

\subsection{Multi-Task Training}
To further elevate the LLM's performance during the second pass, we introduce an enhancement by transforming the model into a multi-task architecture. Alongside the primary task of next-token prediction, we incorporate a secondary task: next-token class prediction. Notably, this distinction sets it apart from conventional Named Entity Recognition (NER)~\cite{10097163, mo2023multi, tong2022improving}, which centers on predicting the entity class for the current token.


We take the backbone of the LLM and initiate LoRA~\cite{hu2021lora} weights to it for efficient training.
The input to the model encompasses the biasing list of entities alongside the input hypothesis. After the backbone, we integrate a class tag head to predict the entity class for the next token. The class tag logits will be used to calculate the class loss. Then, we will pick the class with the highest probability as the class for the next token, which will be used to obtain the embedding of the corresponding class. The embedding will then be added with the embedding from the backbone and used to predict the next token. 
While the process may seem straightforward, a key challenge arises due to the non-differentiability of the argmax operation used to select the class with the highest probability. This characteristic prevents the propagation of gradients to preceding layers. To address this, we introduce the Gumbel softmax trick~\cite{jang2016categorical}. This trick reparameterizes the softmax weights, transforming them into a one-hot distribution. With this reparameterization, the argmax operation becomes differentiable, allowing for the joint training of the entire neural network model.
The final loss can be written as,
\begin{equation}
    L = \alpha L_{token} + (1 - \alpha) L_{class},
\end{equation}
where $L_{token}$ and $L_{class}$ are the cross entropy loss for token and class tag prediction, respectively. $\alpha$ is the task weight and is empirically set to $0.7$ in our experiments.

\subsection{Dynamic Prompting}
Generally, users' biasing lists can be long and contain multiple entity classes, e.g., locations and person names. With such a long list, the LLMs' attention mechanism could be confused, causing a performance drop.
Another problem is that current LLMs have maximum sequence length restrictions; therefore, the length of the biasing list is also limited.

\begin{figure}[t!]
    \centering
    \includegraphics[width=0.38\textwidth]{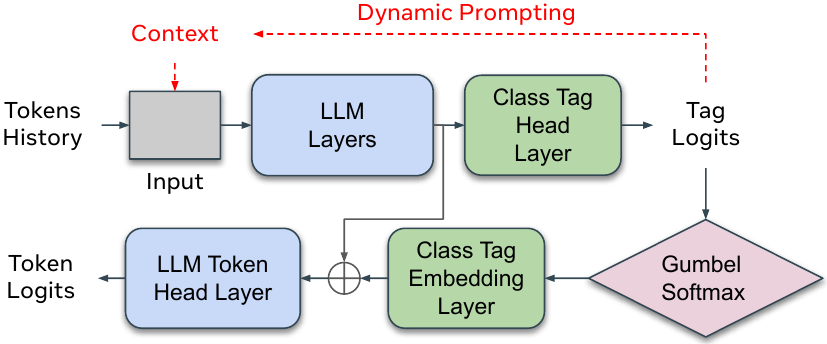}
    \caption{\label{fig:train}
    The flow chart of the proposed multi-task model with dynamic prompting. Red dashed arrows represent the dynamic prompting process. Black solid arrows represent token prediction given the selected prompts.
    }
    \vspace{-0.1in}
\end{figure}

To solve these problems, we propose dynamic prompting, where, instead of having every entity class in the biasing list, we only provide the biasing list corresponding to the next token's entity class. To achieve this, for every token in the hypothesis, we first obtain the class prediction results from the class head by only feeding the hypothesis to the model, and then, we select the corresponding entities in the biasing list and combine them with the hypothesis as input for the LLMs for the final token prediction. This not only reduces the length of the sequence but also reduces the number of entities the model needs to attend to, consequently improving the model's performance. A visual representation of the multi-task model's structure is depicted in Figure~\ref{fig:train}.

With dynamic prompting, the log-likelihood for the sentence becomes a conditional log-likelihood, which can be written as follows,
\begin{align}
    \log P(w_T, ..., w_0|C)
    &=\sum_{t=1}^T \log P(w_t|h_w, c(h_w)),
\end{align}
where $w_t$ is the token at position $t$, $h_w$ is token history, $C$ is the contextual information for all classes, and $c(\cdot)$ is the dynamic prompting function, which selects the contextual information for one specific class given the token history.

\section{Performance Evaluation}

\subsection{Datasets and Comparison Models}

To evaluate the proposed method, we performed extensive experiments on two datasets. 
The first one is an in-house dataset containing calling, messaging, and dictation utterances (denoted as CMD), which are short and contain massive entities. For this dataset, the training data has 460,000 samples, and the testing data contains 14,000 samples.
We use an in-house audio model, with 13 transformer layers as the encoder and 1 Long-Short Term Memory~(LSTM) model as the decoder, for the first-pass score calculation.
Since there is no available public contextual biasing dataset to the best of our knowledge, we reformatted the SLUE-VoxPopuli~\cite{shon2022slue} dataset for our evaluation as the second dataset. SLUE-VoxPopuli can be used for NER, and It contains the ground truth entities, which are further filtered by us only to keep person names, locations, and organizations.
However, there are no ground truth entities in the test dataset in SLUE-Voxpopuli; therefore, we used the validation dataset as the test set in our experiments.
The training set contains 5,000 samples, and the test set contains 1,753 samples. We used wav2vec2~\cite{baevski2020wav2vec} as the audio model, which is augmented by a 4-gram language model during the first pass\footnote{https://huggingface.co/patrickvonplaten/wav2vec2-base-960h-4-gram}.
We train the model on the ground truths texts (i.e., the references) and evaluate it by feeding the hypothesis to it to obtain the second pass scores. Then, the utterance with the highest score will be used as the final hypothesis for calculating the WER. When training the multi-task model, since the entity class is highly unbalanced, we assign weight $0.33$ to all three entity classes and assign $0.01$ to non-entity tokens.

For comparison,
since we are using LLMs, it would be unfair to compare against non-LLM-based methods~\cite{pundak2018deep, chen2019joint}; therefore, we only present the comparison of different variations of the prompting method and multi-task fine-tuning method.
We select LLaMA~\cite{touvron2023llama} and LLaMA 2~\cite{touvron2023llama2} as the baseline models and, for each scenario in the experiment, we select the LLaMA model that has the best results. We also include the evaluation results on RoBERTa~\cite{liu2019roberta} to see the performance of BERT-based model. We calculate the Pseudo Log-Likelihood (PLL) by iteratively masking each token in the input sentence and adding up the obtained log-probability~\cite{xu2022rescorebert, salazar2020masked}.
Since RoBERTa has a maximum sequence length of 512, it is not capable of handling few-shot examples, and thus, we do not present the few-shot learning results for RoBERTa.
Additionally, we also calculate the WER of an oracle, which is the lowest possible WER given the n-best hypothesis.

\begin{table}[!t]
\centering
\caption{\label{tab:no-fine-tune} Second-pass WER evaluation results on different variations of un-fine-tuned LLMs.}
\vspace{-.1in}
\begin{tabular}{l|c|c}
\toprule
\multicolumn{1}{c|}{\textbf{Model}}       & \textbf{CMD}            & \textbf{SLUE}            \\ \hline
First Pass Baseline  & 7.10           & 20.03           \\
\hline
Oracle      & 4.37 (38.5\%) & 16.84 (16.0\%) \\
\hline
LLaMA & 7.07 (0.4\%)  & 18.35 (8.4\%)  \\
\quad+Biasing in Inference & 5.92 (16.6\%) & 18.11 (9.6\%)  \\
\qquad+Few-shot Examples & \textbf{5.84 (17.8\%)} & \textbf{18.10 (9.6\%)}  \\
\hline
RoBERTa & 7.10 (0\%) & 19.60 (2.1\%) \\
\quad + Biasing in Inference & 7.10 (0\%) & 19.73 (1.5\%)  \\
\bottomrule
\end{tabular}
\end{table}


\begin{table}[!t]
\centering
\caption{\label{tab:fine-tune} Second-pass WER evaluation results on different variations of fine-tuned LLMs.}
\vspace{-.1in}
\begin{tabular}{l|c|c}
\toprule
\multicolumn{1}{c|}{\textbf{Model}}      & \textbf{CMD}            & \textbf{SLUE}            \\ \hline
Fine-tuned LLaMA & 7.03 (1.0\%)  & 18.00 (10.1\%) \\
\quad+Biasing in Inference & 5.93 (16.5\%) & 17.98 (10.2\%) \\
\qquad+Biasing in Training & 5.75 (19.0\%) & 18.02 (10.0\%) \\
\hline
Fine-tuned RoBERTa & 7.06 (0.6\%) & 19.11 (4.6\%) \\
\quad+Biasing in Inference & 7.07 (0.4\%) & 19.15 (4.4\%) \\
\qquad+Basing in Training & 7.06 (0.6\%) & 19.16 (4.3\%) \\
\hline
Multi-Task LLaMA & 6.99 (1.6\%)  & 17.95 (10.4\%) \\
\quad+Biasing in Both & 5.71 (19.6\%) & 17.85 (10.9\%) \\
\qquad+Dynamic Prompting & \textbf{5.68 (20.0\%)} & \textbf{17.77 (11.3\%)} \\
\bottomrule
\end{tabular}
\vspace{-0.1in}
\end{table}

\vspace{-.1in}
\subsection{Evaluation Results}
We summarized the evaluation results in Table~\ref{tab:no-fine-tune} and \ref{tab:fine-tune}. Table~\ref{tab:no-fine-tune} contains the results from un-fine-tuned models, while Table~\ref{tab:fine-tune} contains the results from fine-tuned models. \textit{The stacked model component name means it is an add-on of the model in the previous row.}

\begin{figure}[t!]
    \vspace{0.1in}
    \centering 
    \begin{subfigure}[b]{0.49\textwidth}
        \centering 
        \includegraphics[width=1\textwidth]{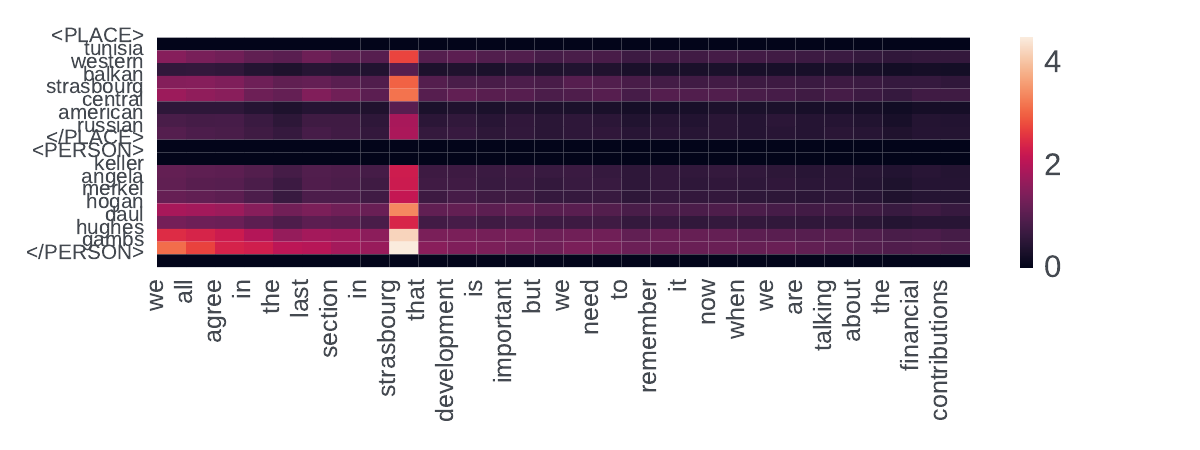}
        \caption{\label{fig:attn0} Attention weights of the first attention layer.}
    \end{subfigure}\hfil
    \begin{subfigure}[b]{0.49\textwidth}
        \centering 
        \includegraphics[width=1\textwidth]{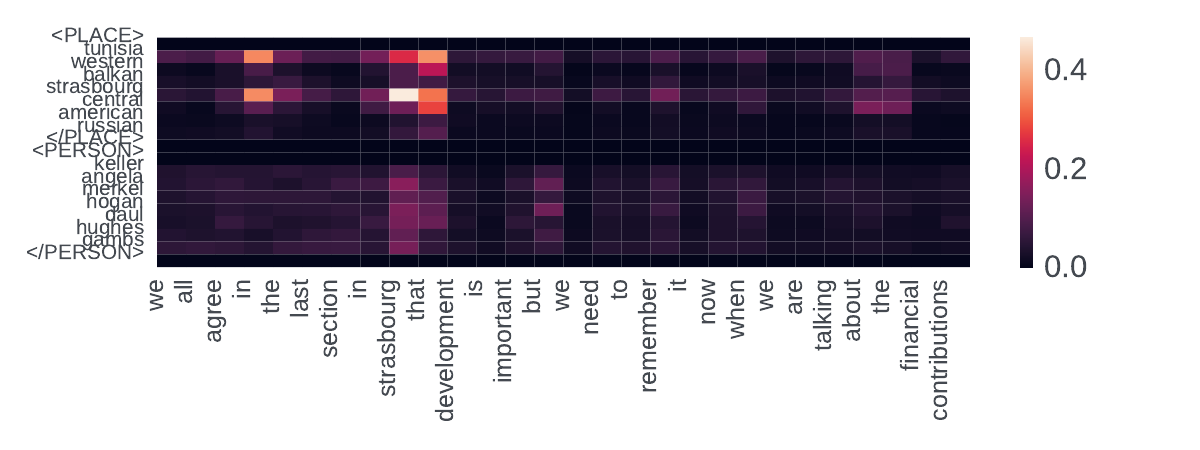}
        \caption{\label{fig:attn24} Attention weights of the $24$th attention layer.}
    \end{subfigure}\hfil
    \begin{subfigure}[b]{0.49\textwidth}
        \centering
        \includegraphics[width=1\textwidth]{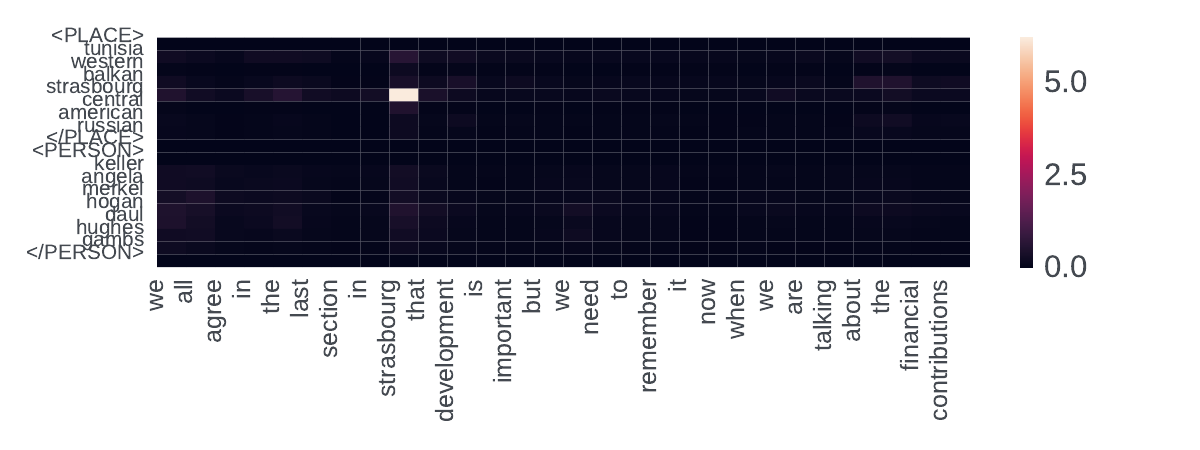}
        \caption{\label{fig:attn32} Attention weights of the last attention layer.}
    \end{subfigure}
    \caption{\label{fig:visual} The attention weights for different layers in the LLaMA model we fine-tuned. The y-axis contains the words in the biasing list, and the x-axis contains the words in the input sentence. ``Strasbourg" is in the input sentence, and the attention layers are able to attend to it in the biasing list.}
    \vspace{-.2in}
\end{figure}

From Table~\ref{tab:no-fine-tune}, we can observe that the biasing list can improve the model's performance significantly. Furthermore, few-shot learning can bring a small improvement because the model can comprehend and learn from them. We can also observe that the entity list is not helping RoBERTa to improve the WER; the reason is that RoBERTa is trained purely as a language model, and it is not good at working with prompts.
In Table~\ref{tab:fine-tune}, we can observe that simply fine-tuning the model on the corpus does not bring an obvious improvement because WER is not like perplexity, and well-trained on a corpus does not necessarily mean it can do well in the second pass rescoring.
We can also observe that fine-tuned models can outperform few-shot learning results in Table~\ref{tab:no-fine-tune},
which is within expectation. 
The results in this table about RoBERTa further prove our claim that it is not suitable for prompt inputs.
Then, we can observe that adding multi-task training can indeed bring an improvement, and we owe this to the training on class tag prediction loss. Finally, we can see that, for fine-tuned models, few-shot learning is not helping; instead, the model performance even deteriorates. 
Another thing worth noting is that the F1 score for the entity class prediction can reach $95\%$, indicating that the model is well-trained for the entity class prediction task.

We also plotted the attention weights for different layers in the multi-task LLaMA model without dynamic prompting in Figure~\ref{fig:visual} with simple and short biasing lists of two classes. We can see that in the beginning (Figure~\ref{fig:attn0}), the attention weights are evenly distributed, and then at the $24$th layer (Figure~\ref{fig:attn24}), the model learns to bias on the list of places. Finally, in the last layer, it assigns the highest weight to the correct place name ``Strasbourg" (Figure~\ref{fig:attn32}).
From the visualization results, we can see that the model is able to learn to bias on the biasing list in the prompts.


\begin{figure}[t!]
    \centering
    \includegraphics[width=0.35\textwidth]{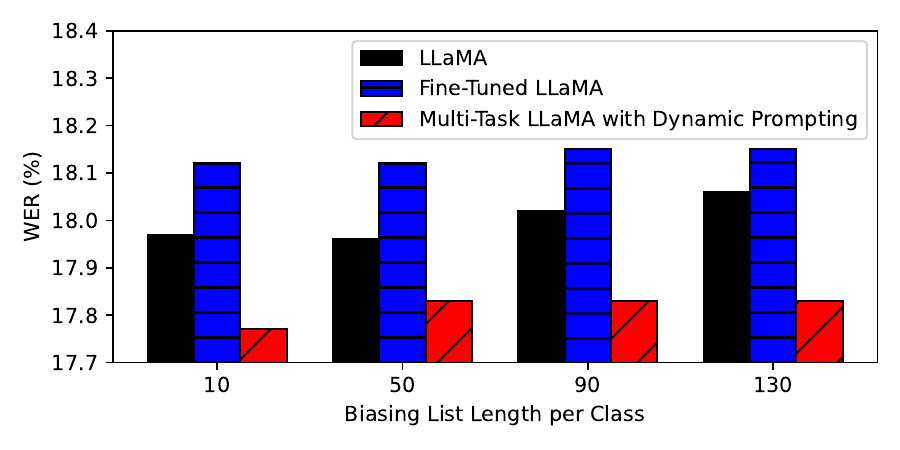}
    \caption{\label{fig:var_len}
    Model performance when the length of the biasing list varies. We have three classes of entities in total.
    }
    \vspace{-.1in}
\end{figure}

Furthermore, in Figure~\ref{fig:var_len}, we vary the length of the biasing list to see the performance difference of the models. We can see that the multi-task model with dynamic prompting is not affected much by the length change because it can filter unrelated entity classes.

\vspace{-.1in}
\subsection{Ablation Study}
The results have indicated that our contextual biasing method can bring a significant improvement in the second pass. However, what if there are no ground truth entities in the biasing list? How will the models' performances be influenced? To answer this question, we perform the ablation study on the appearance of ground truth in the biasing list. The results are summarized in Table~\ref{tab:ablation}.
We randomly select entities to form CMD-NGT, which does not contain the ground truth entity in the utterance. Then, we replace some of the entities in the previous random list with the ground truth ones to form the CMD-GT dataset. From Table~\ref{tab:ablation}, we can observe that the WER can be improved significantly with ground truth. Without ground truth entities, the improvement is not that obvious. However, we emphasize that, even without ground truth entities, the results are still better than the first-pass results. This is because the LLaMA model has been trained on many corpora and itself is a powerful language model. Therefore, we can claim that it is safe to use our method even if there is no ground truth in the biasing list.


\begin{table}[!t]
\centering
\caption{\label{tab:ablation} Ablation study on the impact of the appearance of ground truth entities in the biasing list in terms of WER.}
\begin{tabular}{l|l|l}
\toprule
\textbf{Model}       & \textbf{CMD-GT}            & \textbf{CMD-NGT}            \\ \hline
LLaMA & - & -  \\
\quad+Biasing in Inference  & 5.92 (16.6\%)   & 6.91 (2.7\%) \\
\hline
Fine-tuned LLaMA & - & - \\
\quad+Biasing in Inference & 5.93 (16.5\%) & 6.95 (2.1\%) \\
\qquad+Biasing in Training & 5.75 (19.0\%) & 7.02 (1.1\%) \\
\qquad\quad+Dynamic Prompting & 5.71 (19.6\%) & 7.02 (1.1\%) \\
\bottomrule
\end{tabular}
\vspace{-.15in}
\end{table}

\vspace{-.1in}
\section{Conclusion}
\vspace{-.1in}

To the best of our knowledge, this paper is the first paper for contextual biasing with LLMs. We provide the contextual information in the prompts, leveraging the ability of the LLMs to work with prompts.
We also propose multi-task training, augmenting LLMs with class tag prediction. Both methods are rigorously evaluated on in-house and SLUE-Voxpopuli datasets, demonstrating significant improvements: few-shot learning yields 17.8\% and 9.6\% enhancements, and multi-task fine-tuning achieves 20.0\% and 11.3\% improvements. Ablation studies confirm the role of ground truth in biasing, yet our model remains robust even without it.

\vfill\pagebreak

\small
\bibliographystyle{IEEEbib}
\bibliography{ref}

\begin{thebibliography}{10}

\bibitem{zhao2019shallow}
Ding Zhao, Tara~N Sainath, David Rybach, Pat Rondon, Deepti Bhatia, Bo~Li, and
  Ruoming Pang,
\newblock ``Shallow-fusion end-to-end contextual biasing.,''
\newblock in {\em Interspeech}, 2019, pp. 1418--1422.

\bibitem{le2021deep}
Duc Le, Gil Keren, Julian Chan, Jay Mahadeokar, Christian Fuegen, and Michael~L
  Seltzer,
\newblock ``Deep shallow fusion for rnn-t personalization,''
\newblock in {\em 2021 IEEE Spoken Language Technology Workshop (SLT)}. IEEE,
  2021, pp. 251--257.

\bibitem{le21_interspeech}
Duc Le, Mahaveer Jain, Gil Keren, Suyoun Kim, Yangyang Shi, Jay Mahadeokar,
  Julian Chan, Yuan Shangguan, Christian Fuegen, Ozlem Kalinli, Yatharth Saraf,
  and Michael~L. Seltzer,
\newblock ``{Contextualized Streaming End-to-End Speech Recognition with
  Trie-Based Deep Biasing and Shallow Fusion},''
\newblock in {\em Proc. Interspeech 2021}, 2021, pp. 1772--1776.

\bibitem{ma2023adaptive}
Yingyi Ma, Zhe Liu, and Xuedong Zhang,
\newblock ``Adaptive multi-corpora language model training for speech
  recognition,''
\newblock in {\em ICASSP 2023-2023 IEEE International Conference on Acoustics,
  Speech and Signal Processing (ICASSP)}, 2023.

\bibitem{7178957}
Petar Aleksic, Cyril Allauzen, David Elson, Aleksandar Kracun, Diego~Melendo
  Casado, and Pedro~J. Moreno,
\newblock ``Improved recognition of contact names in voice commands,''
\newblock in {\em 2015 IEEE International Conference on Acoustics, Speech and
  Signal Processing (ICASSP)}, 2015, pp. 5172--5175.

\bibitem{1326104}
S.R. Maskey, M.~Bacchiani, B.~Roark, and R.~Sproat,
\newblock ``Improved name recognition with meta-data dependent name networks,''
\newblock in {\em 2004 IEEE International Conference on Acoustics, Speech, and
  Signal Processing}, 2004, vol.~1, pp. I--789.

\bibitem{brown1992class}
Peter~F Brown, Vincent~J Della~Pietra, Peter~V Desouza, Jennifer~C Lai, and
  Robert~L Mercer,
\newblock ``Class-based n-gram models of natural language,''
\newblock {\em Computational linguistics}, vol. 18, no. 4, pp. 467--480, 1992.

\bibitem{9747573}
Antoine Bruguier, Duc Le, Rohit Prabhavalkar, Dangna Li, Zhe Liu, Bo~Wang, Eun
  Chang, Fuchun Peng, Ozlem Kalinli, and Michael~L. Seltzer,
\newblock ``Neural-fst class language model for end-to-end speech
  recognition,''
\newblock in {\em ICASSP 2022 - 2022 IEEE International Conference on
  Acoustics, Speech and Signal Processing (ICASSP)}, 2022, pp. 6107--6111.

\bibitem{9383560}
Duc Le, Gil Keren, Julian Chan, Jay Mahadeokar, Christian Fuegen, and
  Michael~L. Seltzer,
\newblock ``Deep shallow fusion for rnn-t personalization,''
\newblock in {\em 2021 IEEE Spoken Language Technology Workshop (SLT)}, 2021,
  pp. 251--257.

\bibitem{9687895}
Feng-Ju Chang, Jing Liu, Martin Radfar, Athanasios Mouchtaris, Maurizio
  Omologo, Ariya Rastrow, and Siegfried Kunzmann,
\newblock ``Context-aware transformer transducer for speech recognition,''
\newblock in {\em 2021 IEEE Automatic Speech Recognition and Understanding
  Workshop (ASRU)}, 2021, pp. 503--510.

\bibitem{salemi2023lamp}
Alireza Salemi, Sheshera Mysore, Michael Bendersky, and Hamed Zamani,
\newblock ``Lamp: When large language models meet personalization,''
\newblock {\em arXiv preprint arXiv:2304.11406}, 2023.

\bibitem{fu2023robust}
Xuandi Fu, Kanthashree~Mysore Sathyendra, Ankur Gandhe, Jing Liu, Grant~P
  Strimel, Ross McGowan, and Athanasios Mouchtaris,
\newblock ``Robust acoustic and semantic contextual biasing in neural
  transducers for speech recognition,''
\newblock in {\em ICASSP 2023-2023 IEEE International Conference on Acoustics,
  Speech and Signal Processing (ICASSP)}. IEEE, 2023, pp. 1--5.

\bibitem{pundak2018deep}
Golan Pundak, Tara~N Sainath, Rohit Prabhavalkar, Anjuli Kannan, and Ding Zhao,
\newblock ``Deep context: end-to-end contextual speech recognition,''
\newblock in {\em 2018 IEEE spoken language technology workshop (SLT)}. IEEE,
  2018, pp. 418--425.

\bibitem{chen2019joint}
Zhehuai Chen, Mahaveer Jain, Yongqiang Wang, Michael~L Seltzer, and Christian
  Fuegen,
\newblock ``Joint grapheme and phoneme embeddings for contextual end-to-end
  asr.,''
\newblock in {\em Interspeech}, 2019, pp. 3490--3494.

\bibitem{jain2020contextual}
Mahaveer Jain, Gil Keren, Jay Mahadeokar, Geoffrey Zweig, Florian Metze, and
  Yatharth Saraf,
\newblock ``Contextual rnn-t for open domain asr,''
\newblock {\em Proc. Interspeech 2020}, pp. 11--15, 2020.

\bibitem{fathullah2023prompting}
Yassir Fathullah, Chunyang Wu, Egor Lakomkin, Junteng Jia, Yuan Shangguan,
  Ke~Li, Jinxi Guo, Wenhan Xiong, Jay Mahadeokar, Ozlem Kalinli, et~al.,
\newblock ``Prompting large language models with speech recognition
  abilities,''
\newblock {\em arXiv preprint arXiv:2307.11795}, 2023.

\bibitem{10096429}
Tongzhou Chen, Cyril Allauzen, Yinghui Huang, Daniel Park, David Rybach,
  W.~Ronny Huang, Rodrigo Cabrera, Kartik Audhkhasi, Bhuvana Ramabhadran,
  Pedro~J. Moreno, and Michael Riley,
\newblock ``Large-scale language model rescoring on long-form data,''
\newblock in {\em ICASSP 2023 - 2023 IEEE International Conference on
  Acoustics, Speech and Signal Processing (ICASSP)}, 2023, pp. 1--5.

\bibitem{udagawa2022effect}
Takuma Udagawa, Masayuki Suzuki, Gakuto Kurata, Nobuyasu Itoh, and George Saon,
\newblock ``Effect and analysis of large-scale language model rescoring on
  competitive asr systems,''
\newblock {\em arXiv preprint arXiv:2204.00212}, 2022.

\bibitem{openai_2022}
OpenAI,
\newblock ``Chatgpt: Optimizing language models for dialogue,'' Feb 2022.

\bibitem{touvron2023llama}
Hugo Touvron, Thibaut Lavril, Gautier Izacard, Xavier Martinet, Marie-Anne
  Lachaux, Timoth{\'e}e Lacroix, Baptiste Rozi{\`e}re, Naman Goyal, Eric
  Hambro, Faisal Azhar, et~al.,
\newblock ``Llama: Open and efficient foundation language models,''
\newblock {\em arXiv preprint arXiv:2302.13971}, 2023.

\bibitem{10097163}
Yawen Yang, Xuming Hu, Fukun Ma, Shu’Ang Li, Aiwei Liu, Lijie Wen, and
  Philip~S. Yu,
\newblock ``Gaussian prior reinforcement learning for nested named entity
  recognition,''
\newblock in {\em ICASSP 2023 - 2023 IEEE International Conference on
  Acoustics, Speech and Signal Processing (ICASSP)}, 2023, pp. 1--5.

\bibitem{mo2023multi}
Ying Mo, Hongyin Tang, Jiahao Liu, Qifan Wang, Zenglin Xu, Jingang Wang, Wei
  Wu, and Zhoujun Li,
\newblock ``Multi-task transformer with relation-attention and type-attention
  for named entity recognition,''
\newblock in {\em ICASSP 2023-2023 IEEE International Conference on Acoustics,
  Speech and Signal Processing (ICASSP)}. IEEE, 2023, pp. 1--5.

\bibitem{tong2022improving}
Yiqi Tong, Fuzhen Zhuang, Deqing Wang, Haochao Ying, and Binling Wang,
\newblock ``Improving biomedical named entity recognition with a unified
  multi-task mrc framework,''
\newblock in {\em ICASSP 2022-2022 IEEE International Conference on Acoustics,
  Speech and Signal Processing (ICASSP)}. IEEE, 2022, pp. 8332--8336.

\bibitem{hu2021lora}
Edward~J Hu, Yelong Shen, Phillip Wallis, Zeyuan Allen-Zhu, Yuanzhi Li, Shean
  Wang, Lu~Wang, and Weizhu Chen,
\newblock ``Lora: Low-rank adaptation of large language models,''
\newblock {\em arXiv preprint arXiv:2106.09685}, 2021.

\bibitem{jang2016categorical}
Eric Jang, Shixiang Gu, and Ben Poole,
\newblock ``Categorical reparameterization with gumbel-softmax,''
\newblock {\em arXiv preprint arXiv:1611.01144}, 2016.

\bibitem{shon2022slue}
Suwon Shon, Ankita Pasad, Felix Wu, Pablo Brusco, Yoav Artzi, Karen Livescu,
  and Kyu~J Han,
\newblock ``Slue: New benchmark tasks for spoken language understanding
  evaluation on natural speech,''
\newblock in {\em ICASSP 2022-2022 IEEE International Conference on Acoustics,
  Speech and Signal Processing (ICASSP)}. IEEE, 2022, pp. 7927--7931.

\bibitem{baevski2020wav2vec}
Alexei Baevski, Yuhao Zhou, Abdelrahman Mohamed, and Michael Auli,
\newblock ``wav2vec 2.0: A framework for self-supervised learning of speech
  representations,''
\newblock {\em Advances in neural information processing systems}, vol. 33, pp.
  12449--12460, 2020.

\bibitem{touvron2023llama2}
Hugo Touvron, Louis Martin, Kevin Stone, Peter Albert, Amjad Almahairi, Yasmine
  Babaei, Nikolay Bashlykov, Soumya Batra, Prajjwal Bhargava, Shruti Bhosale,
  et~al.,
\newblock ``Llama 2: Open foundation and fine-tuned chat models,''
\newblock {\em arXiv preprint arXiv:2307.09288}, 2023.

\bibitem{liu2019roberta}
Yinhan Liu, Myle Ott, Naman Goyal, Jingfei Du, Mandar Joshi, Danqi Chen, Omer
  Levy, Mike Lewis, Luke Zettlemoyer, and Veselin Stoyanov,
\newblock ``Roberta: A robustly optimized bert pretraining approach,''
\newblock {\em arXiv preprint arXiv:1907.11692}, 2019.

\bibitem{xu2022rescorebert}
Liyan Xu, Yile Gu, Jari Kolehmainen, Haidar Khan, Ankur Gandhe, Ariya Rastrow,
  Andreas Stolcke, and Ivan Bulyko,
\newblock ``Rescorebert: Discriminative speech recognition rescoring with
  bert,''
\newblock in {\em ICASSP 2022-2022 IEEE International Conference on Acoustics,
  Speech and Signal Processing (ICASSP)}. IEEE, 2022, pp. 6117--6121.

\bibitem{salazar2020masked}
Julian Salazar, Davis Liang, Toan~Q Nguyen, and Katrin Kirchhoff,
\newblock ``Masked language model scoring,''
\newblock in {\em Proceedings of the 58th Annual Meeting of the Association for
  Computational Linguistics}, 2020, pp. 2699--2712.

\end{thebibliography}

\end{document}